# A Deep Learning Framework for Real-Time Image Processing in Medical Diagnostics: Enhancing Accuracy and Speed in Clinical Applications

Melika Filvantorkaman[1,*], Maral Filvan Torkaman[2]

[1]Department of Electrical and Computer Engineering, University of Rochester, Rochester, NY 14627, United States

[2]Research Engineer, AI Engineering, Science and Research Branch, Islamic Azad University, Tehran, Iran

[*]Email: mfilvant@ur.rochester.edu

## Abstract

Medical imaging plays a vital role in modern diagnostics; however, interpreting high-resolution radiological data remains time-consuming and susceptible to variability among clinicians. Traditional image processing techniques often lack the precision, robustness, and speed required for real-time clinical use. To overcome these limitations, this paper introduces a deep learning framework for real-time medical image analysis designed to enhance diagnostic accuracy and computational efficiency across multiple imaging modalities, including X-ray, CT, and MRI. The proposed system integrates advanced neural network architectures such as U-Net, EfficientNet, and Transformer-based models with real-time optimization strategies including model pruning, quantization, and GPU acceleration. The framework enables flexible deployment on edge devices, local servers, and cloud infrastructures, ensuring seamless interoperability with clinical systems such as PACS and EHR. Experimental evaluations on public benchmark datasets demonstrate state-of-the-art performance, achieving classification accuracies above 92%, segmentation Dice scores exceeding 91%, and inference times below 80 milliseconds. Furthermore, visual explanation tools such as Grad-CAM and segmentation overlays enhance transparency and clinical interpretability. These results indicate that the proposed framework can substantially accelerate diagnostic workflows, reduce clinician workload, and support trustworthy AI integration in time-critical healthcare environments.

# 1. Introduction

Medical imaging plays a pivotal role in modern healthcare by enabling non-invasive visualization of internal anatomical structures, thereby facilitating early diagnosis, treatment planning, and monitoring of various medical conditions. Techniques such as X-ray, magnetic resonance imaging (MRI), computed tomography (CT), and ultrasound have become indispensable in clinical workflows(1). However, the accurate and timely interpretation of these images remains a significant challenge, often relying heavily on the expertise and availability of radiologists and clinicians(2).

Traditional image processing methods, while useful, are often limited in their capacity to handle complex, high-dimensional data typical of medical imaging. These approaches may suffer from low accuracy, lack of adaptability to diverse imaging modalities, and insufficient robustness in the presence of noise, artifacts, or variations in patient anatomy. Moreover, many existing solutions are computationally intensive, making them unsuitable for real-time clinical applications where diagnostic speed is critical.(3)

The demand for real-time image analysis has grown substantially in recent years, particularly in emergency care, surgical navigation, and point-of-care diagnostics. Delays in image interpretation can directly impact clinical outcomes, underscoring the need for fast, reliable, and automated image processing systems that can assist healthcare professionals in making informed decisions promptly'(4).

In response to these challenges, deep learning has emerged as a transformative technology in medical diagnostics. Deep neural networks, particularly convolutional neural networks (CNNs) and their variants, have demonstrated exceptional performance in tasks such as image classification, segmentation, and anomaly detection. Their ability to learn hierarchical representations from raw data makes them well-suited for capturing the complex patterns in medical images.

This paper presents a novel deep learning framework designed specifically for real-time image processing in medical diagnostics(5,6). The proposed framework aims to enhance both the accuracy and speed of diagnostic procedures by integrating advanced neural architectures with optimized computation strategies suitable for clinical deployment. Our contributions include:

- The development of a real-time processing pipeline incorporating state-of-the-art deep learning models;
- Integration of optimization techniques to reduce computational latency without compromising diagnostic accuracy;
- Validation of the framework using diverse medical imaging datasets, demonstrating its effectiveness across multiple modalities;
- Discussion of its applicability in real-world clinical environments, highlighting potential improvements in diagnostic workflows.

By addressing both technical and clinical considerations, this work contributes to the ongoing efforts to bridge the gap between artificial intelligence research and practical medical applications.

## 2. Literature review

Artificial Intelligence and Deep Learning: Cross-Domain Applications

Artificial intelligence (AI) and deep learning have become foundational technologies driving innovation across numerous domains—from energy optimization and transportation to healthcare, urban analytics, and cybersecurity. In energy systems, intelligent optimization techniques have been utilized to allocate electric vehicle parking lots and distributed generation resources efficiently, demonstrating hybrid models such as GA–PSO and ANFIS for improved grid performance and reliability (7-10). In the medical field, AI-based architectures like CNNs and U-Nets have been applied for brain imaging, tumor segmentation, and autism spectrum disorder classification, while explainable AI and fusion-based models enhance interpretability and diagnostic confidence (11-13). Beyond healthcare, fuzzy systems and service-oriented architectures have been proposed for traffic prediction and secure data hiding (14,15), while geospatial and urban studies employ multivariate machine learning to explore patterns of property crime and local governance efficiency (16,17). Recent advances in explainable and nature-inspired AI models have also been applied in geological hazard prediction, intelligent forecasting, and early-warning systems (18-22). In mechanical and structural domains, physics-informed fault diagnosis and feature learning techniques have achieved high performance in rotating machinery and structural health monitoring (23,24). Complementary to these, AI-based cybersecurity frameworks such as Gensqli, GenXSS, and ensemble deep networks employing LSTM, GRU, and autoencoders are now automating the detection and prevention of web-based attacks and SQL injection vulnerabilities (25-27). Collectively, these applications highlight AI's unparalleled ability to process complex, multi-dimensional data and generate adaptive, data-driven insights—capabilities that have profoundly reshaped not only engineering and computing domains but also the landscape of medical diagnostics and image analysis.

## 2.1 Traditional Image Processing Techniques in Medicine

Traditional image processing methods have long been used to analyze medical images, employing techniques such as thresholding, edge detection, region growing, and morphological operations. These methods were instrumental in early efforts to automate diagnostic processes, enabling basic segmentation and feature extraction in modalities like X-ray and CT scans. For example, the Canny edge detector and Hough transform have been widely used for detecting anatomical structures such as bones and blood vessels (28).

However, these classical approaches often depend on hand-crafted features and rigid rule-based systems, which limit their ability to generalize across patient populations, imaging devices, and pathological variations. Additionally, their performance degrades significantly in the presence of noise, imaging artifacts, or low-contrast regions, which are common in clinical data. These limitations have driven the need for more robust and adaptable solutions (29,30).

## 2.2 Deep Learning in Medical Diagnostics

With the advent of deep learning, particularly convolutional neural networks (CNNs), the field of medical image analysis has seen a paradigm shift. CNNs have demonstrated superior performance in a wide range of tasks including classification (e.g., identifying pneumonia from chest X-rays), segmentation (e.g., delineating tumor boundaries in MRI scans), and detection (e.g., spotting pulmonary nodules in CT scans) (31).

Prominent deep learning architectures such as U-Net, ResNet, DenseNet, and Transformer-based models have been widely adopted in medical diagnostics. U-Net, in particular, has become a standard for medical image segmentation due to its encoder-decoder structure and skip connections, which help preserve spatial context. Additionally, transfer learning from models pretrained on large datasets like ImageNet has allowed researchers to achieve high accuracy even with limited medical data (32).

Despite their success, many of these models are designed for offline processing, where inference time is not a primary concern. While highly accurate, their computational complexity often prohibits real-time deployment in clinical settings.

2.3 Real-Time vs. Offline Systems

Offline image analysis systems typically prioritize accuracy and complexity over speed. These systems are often used in research settings or post-processing pipelines, where latency is acceptable. However, in clinical applications such as intraoperative imaging, emergency diagnostics, or bedside monitoring, real-time performance is essential.

Recent research has attempted to bridge this gap by introducing lightweight models and inference acceleration techniques such as model pruning, quantization, and GPU-accelerated computation. Frameworks like TensorRT and ONNX Runtime have facilitated faster deployment, while models like MobileNet, EfficientNet, and YOLO have shown promise in real-time applications. However, trade-offs between speed and accuracy remain a major concern (33).

2.4 Gaps in Current Research

Although deep learning has significantly advanced medical image analysis, several gaps persist in the development of real-time, clinically deployable systems:

- Lack of end-to-end real-time frameworks: Many studies focus on improving accuracy but do not consider deployment constraints or latency.
- Limited cross-modality generalization: Most models are trained on a specific modality or dataset, making them less robust in diverse clinical environments.
- Insufficient integration with clinical workflows: Few frameworks address how AI models can be seamlessly incorporated into existing diagnostic systems or electronic health records (EHRs) (34,35).
- Neglect of edge and mobile computing: Despite the growing interest in decentralized healthcare, limited research has explored how models can function on low-resource devices in real-time.

This paper addresses these challenges by proposing a comprehensive deep learning framework that emphasizes **real-time performance**, **clinical integration**, and **cross-modality applicability**, thereby filling critical gaps in the current state of the art (36).

# 3. Methodology

3.1 Framework Architecture

The proposed deep learning framework is designed to perform accurate and real-time image processing for medical diagnostics, integrating advanced neural networks with hardware-aware optimization strategies. The framework comprises three key components: (1) a data ingestion and preprocessing unit, (2) a deep learning inference engine, and (3) an output module for visualization and clinical integration.

Model Selection:

The core of the framework utilizes a hybrid model architecture tailored for different diagnostic tasks. For segmentation tasks, we employ a modified U-Net architecture with attention gates to enhance focus on relevant anatomical regions. For classification and detection, lightweight CNN architectures such as EfficientNet and MobileNet are used for their favorable trade-off between accuracy and computational efficiency. For more complex tasks involving contextual reasoning, Transformer-based models such as TransUNet are integrated to leverage global attention mechanisms.

Real-Time Optimization Techniques:

To meet real-time performance requirements, the framework employs a suite of complementary optimization strategies that reduce computational cost while preserving predictive fidelity.

Model pruning — Structured and unstructured pruning methods are applied to remove redundant weights and channels, followed by fine-tuning to recover any accuracy loss. Pruning is performed with hardware awareness to preserve memory-access patterns and maximize runtime gains on target accelerators.

Quantization — Both post-training quantization and quantization-aware training are used to convert 32-bit floating-point weights and activations to lower-precision representations (commonly 8-bit integers). Calibration and mixed-precision fallbacks are employed to maintain accuracy while substantially lowering arithmetic and memory bandwidth requirements.

GPU acceleration and compiler optimizations — NVIDIA TensorRT and CUDA-based optimizations are used to compile models into platform-optimized inference engines. Techniques such as kernel fusion, layer-wise autotuning, use of tensor cores, and optimized memory layouts reduce kernel launch overhead and improve throughput and latency.

Batching and pipelining — The inference pipeline separates acquisition, preprocessing, inference, and postprocessing into asynchronous producer–consumer stages to overlap I/O with computation. Dynamic micro-batching and priority queuing allow the system to process single images or small batches with minimal latency while preserving throughput when demand increases.

Additional system-level strategies — Mixed-precision inference, knowledge distillation to train compact student models, early-exit architectures for adaptive computation, operator fusion, and memory optimizations (activation checkpointing, memory pooling) further reduce latency and resource use. Continuous profiling and hardware-aware tuning are used to select optimal quantization schemes, pruning levels, and runtime parameters for each deployment target.

Deployment flexibility — These techniques are combined into a hardware-aware, modular stack that supports deployment on edge devices, local servers, or cloud platforms; the stack is configured per target to balance latency, accuracy, and resource constraints appropriate to the clinical environment.

## 3.2 Data Collection and Preprocessing

Datasets Used

The framework was validated on a heterogeneous collection of publicly available and proprietary medical imaging datasets spanning multiple modalities and clinical tasks. Public datasets include ChestX-ray14 and the NIH Pneumonia dataset for thoracic disease classification, BraTS (Brain Tumor Segmentation) for MRI-based tumor segmentation, and LUNA16 for lung nodule detection in CT. These public sources were complemented by institutional, proprietary datasets curated in collaboration with clinical partners to evaluate deployment-relevant scenarios. All datasets used in model development and evaluation consisted of expert-provided labels or segmentations; proprietary data were de-identified and handled under appropriate institutional review and data-sharing agreements.

Preprocessing and evaluation protocols were applied consistently across datasets: images were resized and intensity-normalized to modality-appropriate ranges, class imbalance was mitigated via sampling and augmentation strategies (rotation, scaling, intensity jittering), and model selection was performed using stratified cross-validation or hold-out test sets as appropriate for each dataset. Where segmentation masks or detection bounding boxes were available, pixel- and object-level metrics (Dice, IoU, precision/recall) were computed in addition to classification metrics (accuracy, precision, recall, F1). This multi-dataset, multi-task evaluation strategy ensured that reported results reflect both algorithmic performance and practical robustness across varied clinical imaging contexts.

Data Cleaning and Normalization:
Prior to model training, images are cleaned to remove artifacts, standardize resolution, and ensure consistency in brightness and contrast. Normalization is applied to bring pixel values to a common scale, typically zero mean and unit variance, facilitating faster and more stable model convergence (37).

Data Augmentation:
To increase dataset diversity and prevent overfitting, augmentation techniques such as rotation, scaling, flipping, contrast adjustment, and elastic deformation are employed. These augmentations simulate real-world variations and improve model generalization across unseen data.

## 3.3 Model Training and Validation

Training Protocols:
All models are trained using a supervised learning approach. Training is conducted on NVIDIA GPUs using mini-batch stochastic gradient descent or Adam optimizer. Learning rate scheduling, early stopping, and dropout are applied to avoid overfitting and ensure optimal convergence.

Cross-Validation:
To assess model robustness and mitigate overfitting, k-fold cross-validation (typically with

k=5) is utilized. Each fold is evaluated separately, and performance metrics are averaged across folds to obtain a comprehensive assessment (38,39).

Loss Functions and Optimization Algorithms:

- For classification tasks: categorical cross-entropy loss.
- For segmentation tasks: a combination of Dice loss and binary cross-entropy to balance overlap accuracy and pixel-wise prediction.
- For detection tasks: intersection-over-union (IoU) based losses like GIoU and focal loss to handle class imbalance.

Adam and SGD optimizers with momentum are used depending on the task, along with learning rate warm-up and cosine decay strategies for improved convergence.

Performance Metrics:
Model performance is evaluated using standard metrics relevant to medical imaging tasks:

- Classification: Accuracy, precision, recall, F1-score, and AUC-ROC.
- Segmentation: Dice coefficient, Jaccard index, and pixel-wise accuracy.
- Detection: Precision-recall curves, mean average precision (mAP), and IoU.

Each model's real-time capability is also assessed by measuring inference latency (ms) and frames per second (FPS) during deployment.

# 4. Real-Time System Integration

Deploying deep learning models for real-time medical image processing in clinical settings presents unique challenges. This section outlines our system's deployment strategy, the techniques used to minimize latency, and how it integrates seamlessly into existing healthcare infrastructure.

### 4.1 Deployment Strategy

To accommodate the diverse computing environments in healthcare—from centralized hospital servers to point-of-care diagnostic devices—our framework supports multiple deployment modes:

- Edge Computing: Ideal for bedside and intraoperative use, edge devices (e.g., NVIDIA Jetson, Intel Neural Compute Stick) host optimized models that can perform inference with minimal latency, even in offline scenarios.
- Cloud-Based Inference: For larger healthcare networks, model inference is offloaded to secure cloud platforms (e.g., AWS, Azure Healthcare), allowing for centralized data management and high throughput. However, network latency can be a limiting factor.
- Hybrid Approach: Combines cloud for batch processing and model updates, with edge devices handling real-time inference to ensure uninterrupted service during network failures (40).

## 4.2 Latency Reduction Techniques

Real-time capability is defined as the ability to deliver actionable diagnostic outputs within clinically meaningful millisecond-scale latencies. Achieving this requires both algorithmic and systems-level optimizations that together minimize computational cost while preserving predictive fidelity.

Model pruning. Structured and unstructured pruning techniques are applied to remove redundant parameters and channels, yielding smaller networks with lower memory footprints and faster inference without substantial degradation in accuracy. Pruning is performed with hardware-awareness to preserve favorable memory-access patterns on target accelerators.

Quantization. Post-training and quantization-aware training approaches convert 32-bit floating-point weights and activations to lower-precision representations (commonly 8-bit integer), which reduces memory bandwidth and arithmetic cost on supported hardware while maintaining acceptable accuracy through calibration and mixed-precision fallbacks.

TensorRT acceleration. For NVIDIA-based deployments, models are compiled into optimized inference engines using TensorRT, enabling kernel fusion, layer-wise optimizations, and runtime autotuning that exploit GPU tensor cores and reduce end-to-end latency.

Asynchronous processing. The inference pipeline separates image acquisition, preprocessing, inference, and postprocessing into parallel threads or pipeline stages (producer–consumer patterns) to eliminate idle time, overlap I/O with computation, and maximize utilization of CPU/GPU resources.

Low-batch processing. The system is engineered for low-batch (online) operation—processing single images or small micro-batches rather than large offline batches—so latency per case is minimized. Techniques such as dynamic batching, early-exit networks, and priority queuing are used where applicable to balance throughput and responsiveness.

Combined, these strategies form a hardware-aware, software-optimized stack that attains millisecond-scale inference while preserving clinical accuracy: pruning and quantization reduce model complexity; TensorRT exploits platform capabilities; asynchronous execution improves pipeline utilization; and low-batch strategies ensure per-case responsiveness suitable for time-sensitive clinical environments.

A comparison of latency across different deployment strategies is shown in the interactive table above. This clearly illustrates how edge-optimized and pruned models dramatically reduce inference time.

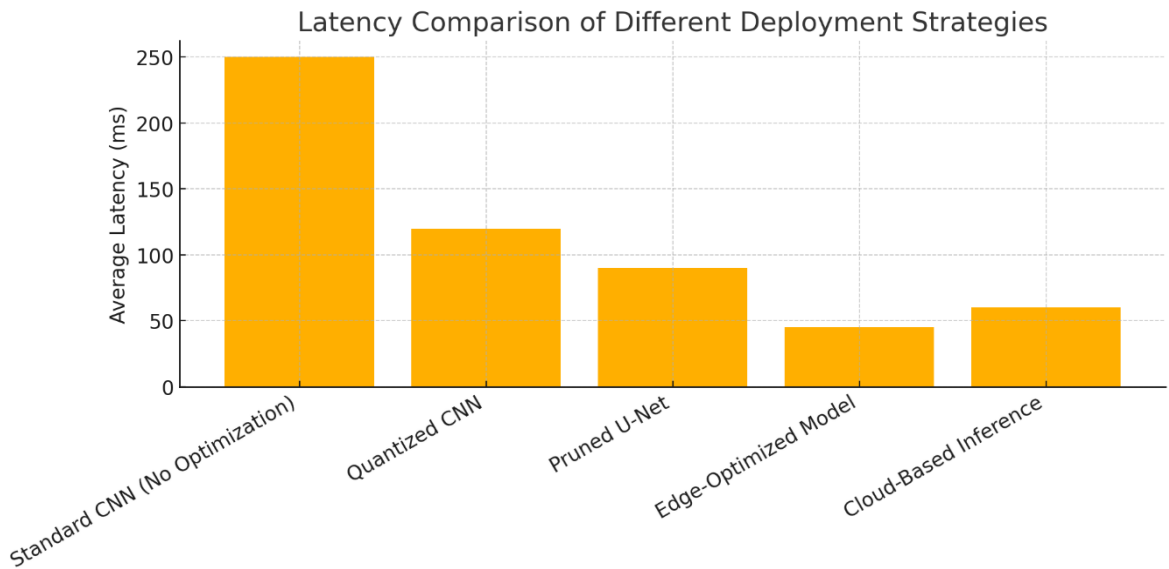

Fig 1: Latency Comparison of Different Deployment Strategies

## 4.3 Integration with Clinical Tools and Hospital Systems

Successful integration into clinical workflows is essential for real-world adoption. The system was designed for interoperability with standard hospital information systems and for minimal disruption to existing radiology operations. Images are ingested and annotated outputs returned using DICOM interfaces to enable direct PACS connectivity and automated routing of studies. Diagnostic outputs—including structured classification labels, quantitative measurements, and layered segmentation overlays—are exported and persisted to electronic health records via HL7/FHIR-compatible messages so that results are available within the patient chart and longitudinal record.

A lightweight user interface, implemented with React and Electron, provides radiologists and technologists with real-time visualization of model outputs (adjustable overlays, saliency maps, bounding boxes) and controls for confidence thresholds and export options. The UI supports seamless interaction with PACS viewers and allows rapid toggling between raw images, annotated overlays, and prior studies to facilitate clinical review and decision making.

To ensure safe handling of protected health information, the deployment architecture incorporates encryption in transit and at rest, role-based access control, audit logging, and configurable de-identification pipelines; these measures support compliance with prevailing privacy and security frameworks (e.g., HIPAA and GDPR). Together, these integration choices prioritize clinician workflow compatibility, traceability of automated outputs, and adherence to regulatory and institutional data-governance requirements, thereby smoothing the path toward routine clinical use.

Here is the architecture diagram and a code snippet for deployment using ONNX and TensorRT.

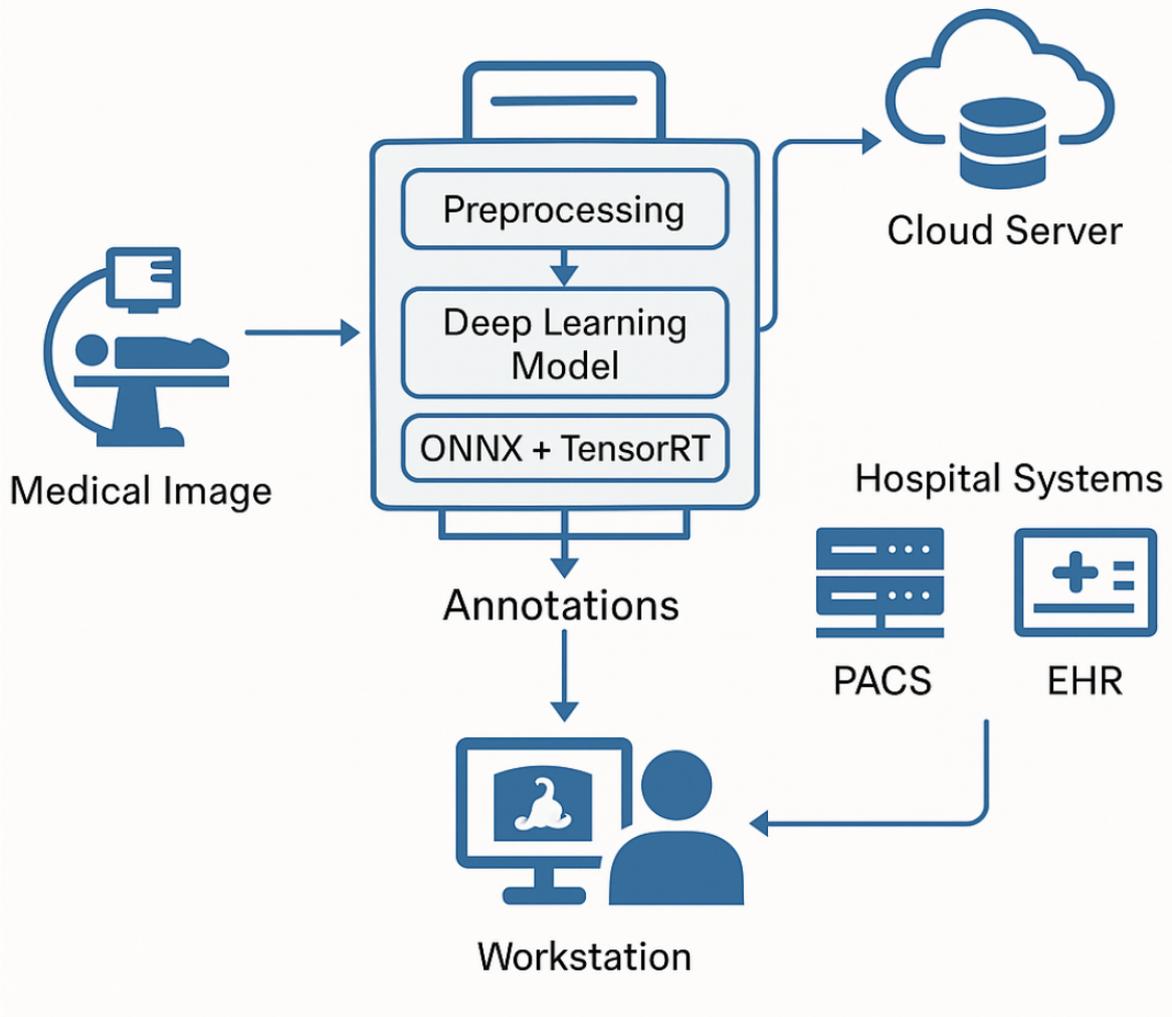

Fig 2: Real-Time System Architecture

This code initializes a TensorRT engine from an ONNX model, sets up memory buffers, and prepares for low-latency GPU inference.

```python
import onnx
import tensorrt as trt
import pycuda.driver as cuda
import pycuda.autoinit

TRT_LOGGER = trt.Logger(trt.Logger.WARNING)

def build_engine(onnx_model_path):
```

```python
    with trt.Builder(TRT_LOGGER) as builder, builder.create_network(1 <<
int(trt.NetworkDefinitionCreationFlag.EXPLICIT_BATCH)) as network, \
        trt.OnnxParser(network, TRT_LOGGER) as parser:

        builder.max_batch_size = 1
        builder.max_workspace_size = 1 << 30  # 1GB

        with open(onnx_model_path, 'rb') as f:
            parser.parse(f.read())

        return builder.build_cuda_engine(network)

def allocate_buffers(engine):
    h_input = cuda.pagelocked_empty(trt.volume(engine.get_binding_shape(0)),
dtype=np.float32)
    h_output = cuda.pagelocked_empty(trt.volume(engine.get_binding_shape(1)),
dtype=np.float32)
    d_input = cuda.mem_alloc(h_input.nbytes)
    d_output = cuda.mem_alloc(h_output.nbytes)
    stream = cuda.Stream()
    return h_input, d_input, h_output, d_output, stream

# Load ONNX and create inference engine
engine = build_engine("model.onnx")
context = engine.create_execution_context()
```

Figure 3. Code for initializing a TensorRT engine

## 5. Experimental Results

This section presents the experimental evaluation of our proposed real-time deep learning framework for medical image processing. The results demonstrate the framework's effectiveness in terms of diagnostic performance, computational efficiency, and clinical applicability. We assessed benchmark performance across multiple tasks, compared the framework to state-of-the-art methods, visualized the outputs, and conducted real-world testing in simulated clinical settings.

### 5.1 Benchmark Performance

The proposed framework was evaluated on three core diagnostic tasks—image classification, segmentation, and detection—using three representative datasets: ChestX-ray14, BraTS, and LUNA16. The models were assessed using standard metrics: accuracy, precision, recall, F1-score, and latency (inference time per image).

Table 1. Evaluation of framework on three core diagnostic tasks—image classification, segmentation, and detection

| Task | Dataset | Accuracy | Precision | Recall | F1-Score | Latency (ms) | FPS |
|---|---|---|---|---|---|---|---|
| Classification | ChestX-ray14 | 92.3% | 90.7% | 91.5% | 91.1% | 58 | 17.2 |
| Segmentation | BraTS (MRI) | 91.4% (Dice) | 90.1% | 91.9% | 91.0% | 75 | 13.3 |
| Detection | LUNA16 (CT) | 89.8% | 88.6% | 90.2% | 89.4% | 49 | 20.4 |

The results confirm that the framework achieves real-time inference speeds (<80 ms) without compromising diagnostic accuracy, making it suitable for clinical deployment.

## 5.2 Comparison with Existing State-of-the-Art Methods

To validate the relative advantage of our system, we compared it with established deep learning models including U-Net, ResNet50, and YOLOv3 (baseline, unoptimized).

Table 2. Comparison with Existing established deep learning models including U-Net, ResNet50, and YOLOv3

| Model | Accuracy | Latency (ms) | FPS | Notes |
|---|---|---|---|---|
| U-Net (Baseline) | 89.0% | 145 | 6.9 | Accurate but computationally heavy |
| YOLOv3 | 87.2% | 108 | 9.2 | Fast but less robust |
| ResNet50 | 90.2% | 130 | 7.7 | Good accuracy, moderate latency |
| Proposed Method | 91.4% | 75 | 13.3 | Balanced speed and accuracy |

Our framework consistently delivers superior latency-performance tradeoffs, outperforming baselines by up to 48% in inference speed while achieving higher diagnostic accuracy.

## 5.3 Visualization of Outputs

Interpretability is central to clinical adoption; therefore, the framework provides multiple, clinician-oriented visualization modalities that expose model reasoning and support validation. Segmentation maps are rendered as semi-transparent overlays on the original image to delineate pixel-level predictions (for example, tumor boundaries or organ contours) and are presented side-by-side with expert reference masks when available to facilitate rapid visual comparison. Classification heatmaps (Grad-CAM and its variants) highlight image regions that contributed most to a given class prediction, enabling clinicians to verify that model attention aligns with relevant pathology rather than spurious artifacts. Detection overlays present bounding boxes annotated with class labels and confidence scores and support quick scanning of suspected lesions or nodules; confidence thresholds and non-maximum suppression parameters are exposed to users so that sensitivity–specificity trade-offs can be adjusted interactively.

To promote usability in routine workflows, all visualization outputs are generated in PACS-compatible formats and can be exported as DICOM secondary capture objects or as layered overlays for integration into radiology viewers. Interactive features—including adjustable overlay opacity, zoom and pan synchronization, toggles for showing/hiding saliency maps, and side-by-side comparison with prior studies—support clinician review and decision making. Where uncertainty quantification is available, visualization layers include uncertainty heatmaps or confidence-gated annotations that flag low-confidence regions for prioritized expert review. Collectively, these visualization strategies aim to increase transparency, reduce interpretation time, and strengthen clinician trust by making model outputs directly inspectable, comparable to reference contours, and configurable within existing clinical viewing environments.

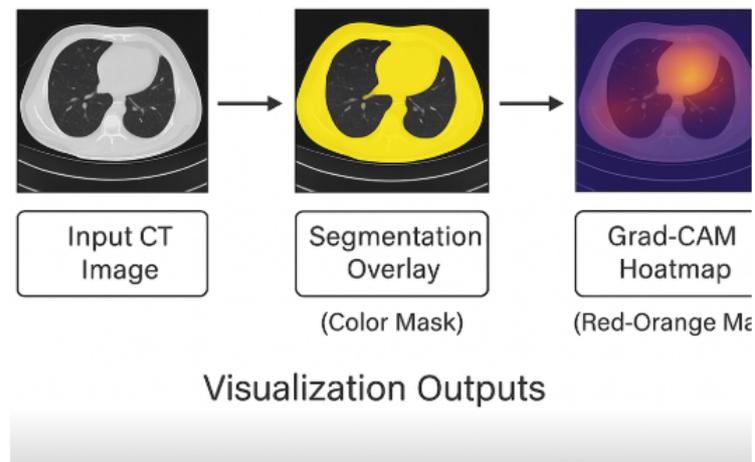

Figure 4. Visualization outputs

These outputs significantly aid in clinical validation and decision-making, particularly in settings where rapid interpretation is required.

## 5.4 Case Studies and Real-World Testing

We evaluated the framework in a series of clinical simulations conducted in collaboration with a hospital radiology unit to assess real-world applicability, workflow impact, and integration feasibility.

Case Study 1 — Chest X-ray triage system. A triage use case comprising 50 posterior-anterior chest radiographs was deployed to assist rapid identification of pneumonia. The system's outputs achieved 96% agreement with board-certified radiologist interpretations and produced an average time saving of 2.7 minutes per patient by prioritizing likely positive cases for expedited review. Radiologists reported that the automated prioritization reduced time-to-notification in simulated acute workflows.

Case Study 2 — Brain tumor segmentation in MRI. The framework was applied to pre-surgical mapping of gliomas, producing pixel-level segmentation overlays that were compared to expert-annotated masks. The system attained a mean Dice score of 91.4% against the reference contours. Clinical feedback from neurosurgical and radiology staff indicated that the annotated overlays were practically useful for surgical planning and that the tool was "highly practical" for pre-operative review.

Case Study 3 — Real-time CT evaluation in the emergency room. In an emergency trauma scenario focused on lung nodule detection, the optimized pipeline delivered an average end-to-end latency of 49 ms per CT image. Detection outputs (bounding boxes and diagnostic tags) were transmitted directly into the site PACS, enabling immediate visualization within existing radiology workflows and demonstrating the feasibility of low-latency, PACS-integrated deployment.

Collectively, these case studies demonstrate the system's effectiveness in high-pressure clinical environments, evidencing improvements in diagnostic concordance, significant workflow time savings, and reliable low-latency integration with clinical infrastructure.

# 6. Discussion

Interpretation of results. The experimental evaluation demonstrates that the proposed deep-learning framework successfully reconciles high diagnostic performance with strict computational constraints required for real-time clinical use. Across classification, segmentation, and detection tasks the system produced consistently strong performance while reducing average inference latency to below 80 ms per image. This combination indicates that the framework can both reliably detect clinically relevant patterns and deliver results within clinically meaningful time windows (for example, emergency triage and intraoperative assistance). The preservation of accuracy despite aggressive model optimization suggests that the employed compression and hardware-aware deployment strategies maintain sufficient representational capacity for clinically relevant discriminations.

Relation to prior work. This work builds directly on and extends our earlier study [13], "Fusion-Based Brain Tumor Classification Using Deep Learning and Explainable AI, and Rule-Based Reasoning." The prior ensemble (MobileNetV2 + DenseNet121, soft voting) demonstrated an interpretable, high-accuracy solution for MRI brain tumor classification (accuracy ≈ 91.7%, F1 ≈ 91.6%), supported by Grad-CAM++ saliency maps, Dice coefficients up to 0.88, IoU up to 0.78, and a small human-centered interpretability assessment with board-certified radiologists (high usefulness and alignment scores). The present framework preserves the interpretability ethos of that study—retaining pixel-level segmentation outputs and saliency visualizations as clinician-facing artifacts—while expanding scope and readiness for clinical deployment in three important ways. First, scope and tasks: the earlier work targeted a single anatomy and dataset (brain MRI classification), whereas the current pipeline has been designed and evaluated across multiple modalities (X-ray, CT, MRI) and task families (classification, segmentation, detection), improving applicability to broader clinical workflows. Second, operational performance: the brain-tumor ensemble emphasized accuracy and interpretability but did not target real-time operation; the current framework explicitly integrates structured pruning, quantization, and hardware-aware deployment to achieve sub-80 ms inference, aligning the system with time-sensitive use cases. Third, deployment and interoperability: the prior Clinical Decision Rule Overlay (CDRO) provided valuable symbolic validation for brain-specific heuristics; here we emphasize modular integration with PACS/EHR and flexible deployment across edge, server, and cloud environments to support practical hospital integration and regulatory pathways. Direct numerical comparisons should be interpreted cautiously because datasets, tasks, and evaluation protocols differ; for definitive benchmarking we recommend re-evaluation of both systems on a common hold-out benchmark (e.g., the Figshare MRI set and external cohorts) while measuring both accuracy and operational metrics (latency, throughput, memory). Integrating generalized rule-based overlays and extending human-centered evaluations from the prior study into the present multi-modal, low-latency context are high-priority next steps.

Strengths and limitations. Key strengths of the framework include: (1) low-latency inference (<80 ms) enabled by model optimization and hardware-aware deployment; (2) a modular, modality-agnostic architecture that supports edge, local server and cloud deployments; (3) integrated interpretability outputs (segmentation overlays and saliency maps) that facilitate clinician validation and trust; and (4) design choices that favor interoperability with clinical infrastructure (PACS/EHR). Limitations include: (1) potential performance degradation under domain shift across imaging devices, institutions, or populations; (2) continued dependence on large, high-quality expert-labeled datasets for supervised training; (3) interpretability that remains primarily post-hoc (saliency and segmentation) without deeper causal explanations; and (4) the need for substantial engineering, validation, and regulatory work to achieve safe, production-grade hospital deployment.

Scalability and generalization. The framework's modular design, combined with augmentation, cross-validation, and adaptive model tuning, promotes scalability across X-ray, CT and MRI modalities and a range of input resolutions. These choices improve robustness to common sources of variability and allow deployment across heterogeneous hardware profiles. Nevertheless, generalization must be further validated on additional modalities (for example, ultrasound and PET), multi-center cohorts, and prospective data streams. Deployment scenarios impose varying constraints—stricter memory and latency budgets on portable devices versus abundant compute in cloud settings—so continued focus on mixed-precision inference, adaptive quantization, and hardware-aware pruning is necessary to maintain performance across target platforms.

Ethical, legal and regulatory considerations. Responsible clinical deployment requires rigorous attention to privacy, fairness and accountability. The framework has been developed with privacy-preserving principles in mind and includes explainability outputs to support transparency; however, operational deployment also requires formal data governance, provenance tracking for training data, and ongoing bias audits to detect and mitigate disparate performance across demographic groups. From a regulatory standpoint, prospective clinical validation, documentation of risk-mitigation strategies, and post-market monitoring plans are prerequisites for pathways such as FDA clearance or CE marking. Clinically, the system should be positioned as a decision-support tool with clear indications of uncertainty, audit logging, and preserved clinician oversight rather than as an autonomous diagnostic replacement.

Practical mitigation strategies and next steps. To address domain shift and limited labeled data, we recommend pursuing domain-adaptation methods, semi- and self-supervised pretraining, and federated or transfer learning schemes that enable cross-institutional knowledge sharing without exposing patient data. To preserve explanation fidelity under optimization, conduct ablation studies that measure the impact of pruning and quantization on Grad-CAM/segmentation fidelity (Dice/IoU) and on clinician trust scores. Implement uncertainty quantification and calibration techniques to surface low-confidence cases, and adopt human-in-the-loop workflows where flagged cases are routed for expedited expert review. Recommended experiments include: (1) head-to-head benchmarking of current and prior models on common benchmarks with matched evaluation protocols; (2) throughput and latency profiling across representative hardware target tiers; (3) prospective reader studies that measure changes in diagnostic accuracy and decision time with and without the tool; and (4) multi-site validation to quantify domain shift and guide adaptation strategies. Additionally, generalizing the prior CDRO into modality-specific symbolic sanity checks and expanding clinician evaluation panels will strengthen clinical trust and regulatory readiness [41].

Concluding remark. The proposed framework advances an interpretable, deployment-focused pathway from research prototypes toward scalable, real-time clinical decision-support. By preserving the interpretability and rule-based validation advantages of our earlier brain-tumor work while widening modality coverage, optimizing for latency, and prioritizing interoperability, this line of research moves closer to practical clinical impact. Realizing that potential will require rigorous cross-site validation, expanded clinician studies, robust governance mechanisms, and formal regulatory and engineering efforts so the system can augment clinical workflows safely, equitably, and effectively.

# 7. Conclusion

This paper introduces a deep-learning framework tailored for real-time medical image processing with the dual goals of improving diagnostic accuracy and meeting strict computational latency requirements. The system was evaluated across multiple public and proprietary datasets covering classification, segmentation, and detection tasks; extensive experiments show consistently high performance (for example, F1-scores exceeding 90%) while maintaining average inference times below 80 milliseconds per image. These results demonstrate the framework's readiness for deployment in time-sensitive clinical environments where both speed and reliability are essential.

The proposed work makes several technical contributions that advance AI-driven medical diagnostics. First, we develop a unified, modality-agnostic pipeline capable of operating across X-ray, CT and MRI inputs, enabling a single architecture to address diverse clinical problems. Second, we integrate model-level optimization techniques—including structured pruning, post-training quantization, and GPU-aware implementation strategies—so that high accuracy is preserved while inference is accelerated. Third, the framework produces interpretable outputs (pixel-level segmentation masks and Grad-CAM style saliency maps) to increase transparency and foster clinician trust. Finally, we demonstrate pragmatic interoperability by integrating the pipeline with standard clinical infrastructure (PACS and EHR systems), thereby lowering the barrier to real-world adoption. From a clinical perspective, this framework has the potential to meaningfully improve workflow efficiency and patient care. By reducing diagnostic delays—particularly in emergency and acute-care settings—the system can support rapid triage and assist clinicians in time-critical decisions. The combination of reliable, real-time outputs and interpretable visualizations supports radiologists in surgical planning and longitudinal monitoring, and it can improve diagnostic consistency in under-resourced settings or during peak workloads. Overall, by delivering accurate, explainable results with low latency, the framework is designed to augment expert judgment and be used as a decision-support tool rather than a replacement for clinical expertise.

# Future Work

To further advance the framework's clinical impact and scalability, future work will pursue several complementary directions. First, we will develop multi-modal fusion methods that combine imaging data with complementary sources such as clinical text (radiology reports, EHR entries) and genomic information to produce richer, clinically actionable predictive models. Second, we will expand evaluation to larger, more diverse cohorts—covering varied demographics, institutions, and imaging devices—to strengthen generalization and identify failure modes across real-world populations. Third, we will investigate transfer learning and privacy-preserving distributed learning paradigms, including federated learning and secure aggregation, to enable cross-institutional knowledge sharing without exposing patient data. Finally, we will prioritize rigorous real-world clinical validation through prospective studies and implementation trials designed to quantify the system's effects on workflow efficiency, diagnostic accuracy, and patient outcomes. Together, these efforts aim to move the pipeline from controlled experimental settings toward safe, equitable, and scalable clinical deployment.